\documentclass[runningheads]{llncs}
\usepackage[T1]{fontenc}
\usepackage{graphicx}
\usepackage{multirow}
\usepackage{subcaption}
\usepackage{xcolor}
\usepackage{amsfonts}
\usepackage{comment}
\usepackage{amsmath}
\usepackage{hyperref}
\usepackage{booktabs}
\usepackage{pifont}

\newcommand{\vecsub}[2]{\mathbf{#1}_{#2}}

\usepackage{graphicx}

\begin{document}
\title{TACENR: Task-Agnostic Contrastive Explanations for Node Representations}

\author{Vasiliki Papanikou\inst{1,2} \and
Evaggelia Pitoura\inst{1,2}}
\authorrunning{V. Papanikou and E. Pitoura}

\institute{
University of Ioannina, Greece \and
Archimedes/Athena Research Center, Greece\\
\email{v.papanikou@athenarc.gr, pitoura@uoi.gr}
}
\maketitle

\begin{abstract}
Graph representation learning has achieved notable success in encoding graph-structured data into latent vector spaces, enabling a wide range of downstream tasks. However, these node representations remain opaque and difficult to interpret. Existing explainability methods primarily focus on supervised settings or on explaining individual representation dimensions, leaving a critical gap in explaining the overall structure of node representations. In this paper, we propose TACENR (Task-Agnostic Contrastive Explanations for Node Representations), a local explanation method that identifies not only attribute features but also proximity and structural ones that contribute the most in the representation space. 
TACENR builds on contrastive learning, through which we learn a similarity function in the representation space, revealing which are the features that play an important role in the representation of a node. 
While our focus is on task-agnostic explanations, TACENR can be applied to supervised scenarios as well. Experimental results demonstrate that proximity and structural features play a significant role in shaping node representations and that our supervised variant performs comparably to existing task-specific approaches in identifying the most impactful features. 


\end{abstract}
\section{Introduction}

Node representation learning plays a significant role in graph-based machine learning, enabling the transformation of graph entities into low-dimensional vector spaces that capture structural and relational information. These representations are widely used across both supervised downstream tasks including node classification and link prediction and unsupervised ones like community detection and similarity search, across diverse fields such as social network analysis and recommendation \cite{fan2019graph}, protein to protein interaction prediction \cite{fout2017protein}, disease-gene association identification \cite{han2019gcn} and other. 

A wide range of models have been developed for learning node representations. Early approaches, such as node2vec \cite{grover2016node2vec}, capture graph structure by simulating random walks or applying matrix factorization techniques. More recently, Graph Neural Networks (GNNs) \cite{kipf2016semi,velickovic2017graph,hamilton2017inductive} have emerged, where node representations are learned by aggregating information from local neighborhoods. However, despite their success, all of these methods lack interpretability, as the dimensions of the resulting representations do not correspond to meaningful structures in the input graph. 

Current explainability techniques in graph learning \cite{yuan2022explainability,nandan2025graphxai} are primarily designed for task-specific scenarios, such as explaining predictions in node classification.
In contrast, the explainability of node representations themselves remains underexplored. Moreover, existing work on the interpretability of node representations focuses on explaining individual representation dimensions \cite{shafi2024generating,piaggesi2024dine,xie2022task}, rather than the representation as a whole.
Node representations encode different types of information: (a) proximity, i.e., neighboring nodes tend to have similar representations, (b) structural similarity, i.e., nodes with similar structural roles share similar representations, and (c) node attribute similarity, i.e., nodes with comparable features remain close in the representation space.
However, existing explanation methods for node representations often limit their scope to a single type of these relationships. 
For instance, some methods focus on identifying subgraphs that most contribute to a specific representation dimension, thereby capturing only proximity \cite{piaggesi2024dine,xie2022task} or only structural features \cite{shafi2024generating}. As a result, these methods fall short of capturing the full information encoded in node representations, which integrates proximity, structure, and attributes.

To address these limitations, we propose TACENR, a local explainability method that offers task-agnostic explanations for node representations. Our method explains the whole node representation by identifying proximity features, structural features, and node attributes that contribute the most to the position of the node in the representation space.
TACENR characterizes the importance of their contribution by building an interpretable model, such as linear regression, to learn the similarity in the representation space between the representation of the node to be explained and the representation of other similar and dissimilar nodes.
Our approach is applicable to both supervised and unsupervised node representations. In the contrastive learning process over similar and dissimilar nodes, we introduce a weighted similarity function that adapts to the type of representation. For supervised settings, the weighting emphasizes dimensions most correlated with the predicted label, while for unsupervised representations, it prioritizes dimensions with the highest variance.

We apply our approach to explain representations produced by node2vec \cite{grover2016node2vec}, role2vec \cite{ahmed2020role} and three GNNs, GCN\cite{kipf2016semi}, GAT\cite{velickovic2017graph}, and GraphSAGE\cite{hamilton2017inductive} on several datasets under both supervised and unsupervised settings. Our experimental results show that proximity, structural features, and node attributes all contribute to the learned node representations but their relative importance varies across representation methods, often reflecting the architecture of the underlying algorithms. For instance, in the case of node2vec, proximity is highly important, indicating that the position of nodes in the graph is prominently encoded in the representations. For GNN models, features related to the local neighborhood, such as the average degree of neighboring nodes and the average neighbor clustering coefficient, are among the most influential, which is expected as GNNs generate node representations by aggregating information from multi-hop neighborhoods.

We also evaluate our supervised variant against other task-specific explanation methods using the AOPC metric \cite{amara2022graphframex}, which measures whether the identified features are truly important for the model and by adding noisy features to assess whether the methods filter them out \cite{huang2022graphlime,ribeiro2016should,duval2021graphsvx}.
Finally, we conduct a comprehensive ablation study to examine how different parameters affect the explanations across datasets and representation models.

The rest of the article is organized as follows: Section \ref{related_work} reviews the related work, Section \ref{TACENR} introduces the TACENR explainer, Section \ref{experiments} presents the experimental results, and Section \ref{conclusion} concludes the paper.

\section{Related Work}
\label{related_work}

The growing adoption of AI models that operate on graph-structured data, driven by their  success across various applications, has generated significant interest in understanding their decision-making processes.
However, most existing explainability methods focus on explaining specific model predictions for downstream tasks, such as node classification or link prediction, with only a few approaches aiming to explain the learned representations.
Given this, we organize related work into two categories of local explanations: task-specific explanations and explanations for node representations. 

\textbf{Task-specific explanations.} Task-specific explanation methods for GNNs \cite{yuan2022explainability}, that operate at the node level, can be broadly categorized into gradient-based \cite{baldassarre2019explainability,pope2019explainability}, decomposition-based \cite{schnake2021higher,schwarzenberg2019layerwise}, perturbation-based \cite{ying2019gnnexplainer,luo2020parameterized,schlichtkrull2020interpreting,duval2021graphsvx,giorgi2025combinex}, and surrogate approaches \cite{huang2022graphlime,zhang2021relex,vu2020pgm}. 
Gradient-based methods (e.g., saliency maps, guided backpropagation), estimate node and edge importance by analyzing the gradients of the model. Decomposition-based methods (e.g., LRP, GNN-LRP) instead propagate prediction scores backward using predefined relevance rules to obtain node and edge importance.
Perturbation based methods study the output variations with respect to different input perturbations. Intuitively, when important input information is retained, the predictions should be similar to the original predictions. One of the most widely used perturbation based methods is GNNExplainer \cite{ying2019gnnexplainer}. This method learns soft edge and feature masks via gradient descent to identify the subgraph and features that maximize mutual information with the original prediction.
GraphSVX \cite{duval2021graphsvx} instead samples binary node and feature coalitions, generates perturbed graphs, evaluates the GNN on each coalition, and trains a weighted linear surrogate to approximate Shapley values.
COMPINEX \cite{giorgi2025combinex}, unlike prior perturbation-based approaches that treat edge and feature masking independently, jointly optimizes perturbations to both graph structure and node attributes. 
Surrogate-based methods operate under the assumption that relationships within a local neighborhood are simpler and can be effectively approximated by an interpretable surrogate model. GraphLIME \cite{huang2022graphlime}, extending LIME \cite{ribeiro2016should}, builds a local neighborhood dataset around the target node, fits an HSIC-Lasso non-linear model, and identifies the most important features contributing to the node prediction.

Our approach focuses on attribution-based explanations derived from proximity, structural, and node attributes, whereas previous methods rely only on edge or attribute importance.
Furthermore, while task-specific methods explain particular predictions and depend on a specific downstream task, our approach is also able to explain node representations independently of any downstream task.
\begin{table*}[ht]
    \caption{Properties of explanations methods for graph-based learning.}
    \centering
    \small
    \begin{tabular}{l|c|c|c}
         & \textbf{Task-agnostic} & \textbf{What explains} & \textbf{Explanation type} \\ \hline
    SA/ Guided-BP \cite{baldassarre2019explainability} & No & label & node/edge imp. \\ 
    CAM \cite{pope2019explainability} & No & label & node imp. \\ 
    GNN LRP \cite{schnake2021higher} & No & label & graph walk imp. \\ 
    COMPINEX \cite{giorgi2025combinex} &No &label& edge/attr. imp.\\
    GNNExplainer \cite{ying2019gnnexplainer} & No & label & node/edge/attr. imp.\\ 
    PGExplainer \cite{luo2020parameterized} & No & label & edge imp. \\ 
    GraphSVX \cite{duval2021graphsvx} & No & label & node/feature imp. \\ 
    ZORRO \cite{funke2022zorro} & No & label & node/attr. imp. \\ 
    GraphLIME \cite{huang2022graphlime} & No & label & attr. imp. \\ 
    RelEx \cite{zhang2021relex} & No & label & node imp. \\ 
    \cite{shafi2024generating} & Yes & dimension & struct./node imp. \\ 
    DINE \cite{piaggesi2024dine} & Yes & dimension & edge imp. \\ 
    TAGE \cite{xie2022task} & Yes & dimension & edge imp. \\ 
    UNR-Explainer \cite{kang2024unr} & Yes & representation & edge imp. \\ 
    TACENR & Yes & representation & proximity/struct./attr. imp.\\ 
    \end{tabular}
    \label{tab:approaches_comparison}
\end{table*}

\textbf{Explanations for node representations.} While many studies address explanations for specific graph tasks, fewer interpret the representation space resulting from unsupervised methods. 
Previous approaches \cite{shafi2024generating,piaggesi2024dine,xie2022task} focus on explaining individual representation dimensions whereas our approach provides explanation for the representation as a whole.
In \cite{shafi2024generating} the authors explain each representation dimension by measuring its similarity to a vector of structural features of the graph, like degree and clustering coefficient, interpreting each similarity score as an indication of how much a particular representation dimension is defined by a given structural feature. Although this method also uses structural features for explanations, it differs from ours by focusing exclusively on individual dimensions. 
DINE \cite{piaggesi2024dine}, provides one explanation per dimension by identifying the substructures each dimension helps to reconstruct. Using an utility function, they quantify the contribution of each dimension to edge reconstruction and interpret higher contributions as greater importance. The most informative edges for each dimension are then used to form subgraphs that serve as explanations. 
TAGE \cite{xie2022task} also explains individual representation dimensions but does so by decomposing the model into a GNN-based representation module and a downstream module. Rather than explaining the full model as a single unit, TAGE introduces separate explainers for each component. The representation explainer, focuses on the representation model by identifying subgraphs that influence the learned node representations and a downstream explainer, which assigns importance scores to these representation dimensions based on the downstream model.
Finally, UNR-Explainer \cite{kang2024unr} explains node representations through counterfactual edge perturbations, identifying subgraphs that influence a node position and neighborhood in the representation space. Unlike our approach, UNR-Explainer provides only edge-attribution explanations, whereas our method explains a node representation by mapping its representation vector to proximity, structural, and attribute features.
Table~\ref{tab:approaches_comparison} summarizes the approaches and compares their main properties.

\section{Explaining Node Representations}
\label{TACENR}
TACENR, generates explanations for node representations by identifying the features, among proximity, structural features and node attributes, that contribute significantly to the representation space.
TACENR learns a similarity function in the representation space among the node being explained and other nodes, in a contrastive manner, capturing how feature differences between nodes translate into similarities in the representation space.

\subsection{TACENR Explainer}

Let $G= (V,E)$ be a graph, where $V$ represents the set of nodes and $E$ the set of edges. Each node $v \in V$ is associated  with an attribute vector $\vecsub{x}{v} \in \mathbb{R}^M$, which encodes its $M$ attributes. We denote by $X \in \mathbb{R}^{|V| \times M}$ the matrix of input features for all nodes. 
We also consider a node representation model \(f:(G, X)\to\ \mathbb{R}^d\), mapping each node \( v \in V \) to a \( d \)-dimensional representation \( \vecsub{z}{v} \in \mathbb{R}^d \). This representation may be derived through either supervised or unsupervised methods that we aim to explain. In some cases, the representation model may take only the graph structure as input  \(f:G\to\ \mathbb{R}^d\) as in node2vec.

We explain a node representation by identifying the features that drive its similarity to other nodes in the representation space. Since node representations aim to capture latent structural and relational patterns in a graph, nodes with similar representations are expected to share common characteristics. Our approach explains a representation by mapping it back to node attributes, proximity features, and structural features, enabling a quantitative assessment of the contribution of each factor to the final representation.

To achieve this, in addition to attribute features, we enrich each node with features capturing its position and role in the graph.
We consider two categories of features: proximity and structural features. 
Proximity features quantify how closely nodes are connected and are derived from distance or diffusion-based metrics, such as shortest-path distance or personalized PageRank.
We denote by $p_{v,u}$ the proximity of node $v$ with respect to another node $u$.
Structural features characterize the functional role of a node in the graph topology and include metrics such as degree, average neighbor degree, and clustering coefficient.
We represent these by a vector $\vecsub{s}{v} \in \mathbb{R}^P$, where $P$ is the dimension of this vector.
The combined feature vector of node $v$ with respect to node $u$ is then
$\vecsub{h}{v,u} \in \mathbb{R}^{M+P+1}$ and is obtained by concatenating its attribute, structural, and proximity features:
\[
\vecsub{h}{v,u} = [\,\vecsub{x}{v} \circ \vecsub{s}{v} \circ p_{v,u}\, ].
\]

To explain the representation of a node $v$ we learn its similarity in the representation space with other nodes. 
Given a similarity measure $\textit{sim}$ between node representations, we define an explanation model
$g: \mathbb{R}^{M+P+1} \to \mathbb{R}$ 
which takes the combined feature vector $\vecsub{h}{v,u}$ as input and predicts the similarity $\textit{sim}(\mathbf{z}_v, \mathbf{z}_u)$ between the representations of nodes $v$ and $u$.
The explanation model is trained by minimizing the following:
\[
\sum_{u \in S_v}
\mathcal{L}\big(
\textit{sim}(\mathbf{z}_v,\mathbf{z}_u),
\, g(\vecsub{h}{v,u})
\big),
\] where $S_v$ is a set of nodes.  
This objective minimizes the discrepancy between the true similarity 
\(\textit{sim}(\mathbf{z}_v,\mathbf{z}_u)\) and the similarity predicted by the explainer \(g(\vecsub{h}{v,u})\).

We learn this function contrastively. 
For each node \(v\), we construct two sets: 
the \textit{affinity set} \(S_v^{+}\), consisting of the \(k_{aff}\) nodes whose 
representations are most similar to \(\vecsub{z}{v}\) according to \textit{sim}, 
and the \textit{divergence set} \(S_v^{-}\), consisting of the \(k_{div}\) nodes 
whose representations are least similar to \(\vecsub{z}{v}\). $S_{v}$ is the union $S_{v}^{+} \cup S_{v}^{-}$ of the two sets.

The computational complexity of TACENR arises from two main components: constructing the contrastive sets $S_{v}^{+}$ and $S_{v}^{-}$, and computing the structural and proximity features. 
To form the affinity and divergence sets, we compute the similarity between the target node $v$ and all other nodes, which requires  $O(|V| \cdot d)$, where 
$d$ is the representation dimension. We then select the $k_{aff}$ most similar and $k_{div}$ least similar nodes using partial selection in $O(|V|  \cdot \log{|S_v|})$, leading to an overall cost of $O\big(|V| \cdot d + |V| \cdot \log{|S_v|}\big)$.

The cost of computing structural and proximity features depends on the type of information they encode.
Basic structural features such as \textit{degree} are obtained in constant time $O(1)$, while first-order neighborhood statistics such as \textit{average neighbor degree} require $O(d_v)$, to iterate over the immediate $d_v$ neighbors of $v$. Features that depend on neighbor interactions, such as \textit{clustering coefficient} or \textit{triangle counts}, require examining pairs of neighbors and incur $O(d_v^2)$. Higher-order aggregated statistics like \textit{average neighbor clustering coefficient} may depend on the maximum degree $\Delta$ giving $O(d_v \cdot \Delta^2)$ in the worst case. For proximity features computing shortest-path distances using BFS costs $O(|V| + |E|)$ while diffusion-based features such as personalized PageRank require iterative propagation with cost
$O(k \cdot |E|)$, where $k$ is the number of iterations. Many of these features can be computed using well-established approximations, significantly reducing runtime on large graphs.

\begin{figure*}[h]
    \centering
    \includegraphics[width=\textwidth]{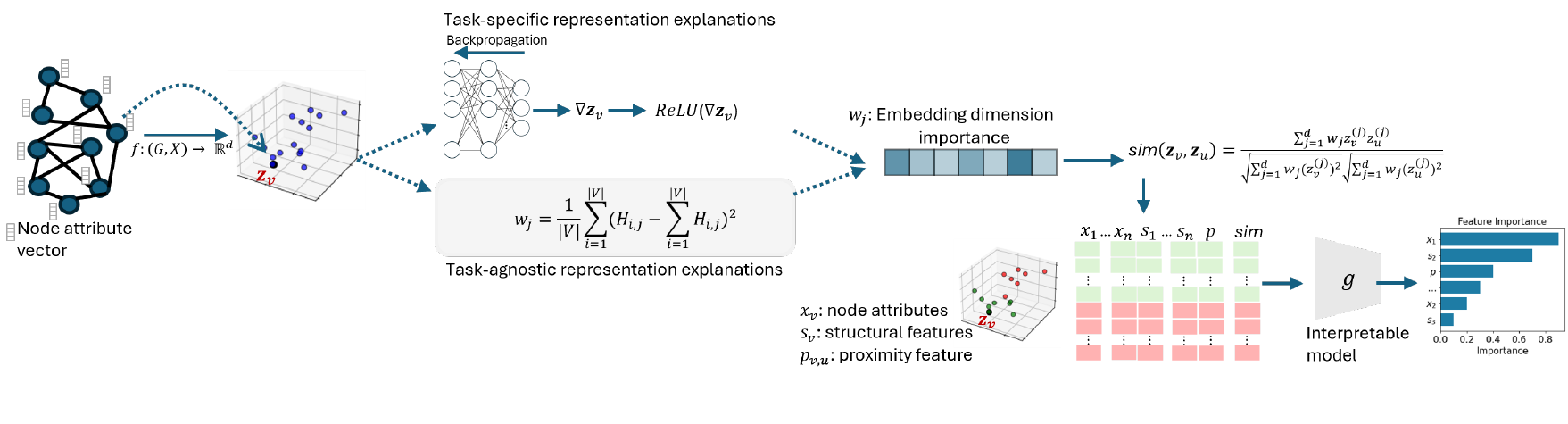}
    \caption{Architecture of the TACENR Explainer.}
    \label{fig:TACENR_architecture}
\end{figure*}

\subsection{Similarity Measure}
We compute the similarity between node representations using cosine similarity.  
To emphasize the representation dimensions that are most informative for interpretation,  
we introduce a weighted variant of cosine similarity:

\[
\textit{sim}(\mathbf{z}_v,\mathbf{z}_u) = 
\frac{
    \sum_{j=1}^{d} w_j \, z_v^{(j)} \, z_u^{(j)}
}{
    \sqrt{ \sum_{j=1}^{d} w_j \, (z_v^{(j)})^2 }\;
    \sqrt{ \sum_{j=1}^{d} w_j \, (z_u^{(j)})^2 }
}.
\]

\noindent \textbf{Weighting for supervised representations.}
In the supervised case, the weights are designed to prioritize the dimensions of representations that are most relevant to the downstream task. 
To achieve this, we use a \textit{gradient-based weighting} approach inspired by gradient-based explainers \cite{baldassarre2019explainability,pope2019explainability}, deriving the weights directly from the trained GNN. This allows us to emphasize the representation dimensions that most strongly drive the prediction for node $v$. 
Let $\mathbf{z}_v \in \mathbb{R}^d$ denote the node representation from the penultimate layer, and let $\mathbf{P}(\mathbf{z}_v) \in [0, 1]^C$ represent the output probability of the model over $C$ classes. We compute the gradient of the predicted class probability with respect to the input representation:
$
\mathbf{g}_v = \nabla_{\mathbf{z}_v} \left( \max_{c \in \{1,\dots,C\}} \mathbf{P}(\mathbf{z}_v)_c \right) \in \mathbb{R}^{d}
$.
To isolate the dimensions that positively contribute to the model confidence, we apply the ReLU function to the gradients. Finally, we normalize the result using the mean absolute magnitude of the gradients. The final weighting vector $\mathbf{w}^{(\text{grad})} \in \mathbb{R}^d$ is defined as:
$
\mathbf{w}^{(\text{grad})} = \frac{\text{ReLU}(\mathbf{g}_v)}{\frac{1}{d} \sum_{j=1}^{d} |\mathbf{g}_{v}^{(j)}|}
$,
where the denominator represents the average gradient magnitude across all dimensions.

\textbf{Weighting for unsupervised representations.}
In the unsupervised setting, defining an appropriate weighting strategy is more challenging, as we aim to identify the most informative dimensions of representations without any task-specific supervision. We therefore compute weights based on the \textit{global variance} of each representation dimension. Not all  dimensions contribute equally to distinguishing nodes. Some dimensions may exhibit little variation across the graph, carrying minimal discriminative information, while others vary significantly, capturing meaningful structural roles or patterns. 
Let $H \in \mathbb{R}^{|V|\times d}$ denote the matrix containing all node representations. To avoid scale bias across representation dimensions, we first normalize the node representations and construct the normalized matrix $H$. 
To prioritize informative dimensions, we compute the weight of a dimension $j$ as the variance of each representation feature across all nodes:
$
w_j = \frac{1}{|V|} \sum_{i=1}^{|V|} \left( H_{i,j} - \mu_j \right)^2$ 
,where $H$ is the representation matrix and $\mu_j = \frac{1}{|V|} \sum_{i=1}^{|V|} H_{i,j}
$.

The overall architecture of the TACENR explainer is illustrated in Figure~\ref{fig:TACENR_architecture}.

\section{Experiments}
\label{experiments}
The goal of our experiments is threefold: (a) examining the behavior of TACENR explanations across various unsupervised representation models, (b) comparing our supervised variant against existing explanation methods, and (c) conducting a comprehensive ablation study.
We use three widely used citation network datasets, Cora \cite{mccallum2000automating}, CiteSeer \cite{giles1998citeseer}, and PubMed \cite{sen2008collective}, a protein–protein interaction dataset PPI \cite{zitnik2017predicting}, and the synthetic BA-Shapes dataset \cite{ying2019gnnexplainer}. 
The node attributes in Cora and CiteSeer are binary word indicators, whereas in PubMed they correspond to TF-IDF values. For the PPI dataset, which contains 121 classes, we focus on the most frequent class for the prediction task.
Detailed statistics for these graphs are provided in Table \ref{tab:graph_stats}. Our full code is available at\footnote{\href{https://github.com/VasilikiPapanikou/TACENR-Task-Agnostic-Contrastive-Explanations-for-Node-Representations}{GitHub code}}.

\begin{table}[htbp]
\centering
\tiny
\caption{Graph statistics and illustration of the BA-Shapes dataset.}
\label{tab:graph_stats}
 
\begin{minipage}{0.63\columnwidth}
    \centering
    \hspace*{0.8cm}
    \begin{tabular}{l|c|c|c|c|c}   
         & \textbf{Cora} & \textbf{CiteSeer} & \textbf{PubMed} & \textbf{PPI} & \textbf{BA-Shapes} \\ \hline
        Number of Nodes ($|V|$)               & 2,485 & 2,120 & 19,717 &3,480  & 700 \\
        Number of Edges ($|E|$)               & 5,069 & 3,679 & 44,324 &54,806  & 1,973 \\
        Avg. Degree                           & 4.07  & 3.47  & 4.49   & 31.49 & 5.66 \\
        Avg. Neighbor Degree                  & 12.48 & 7.58  & 19.27  & 99.12 & 9.33 \\
        Avg. Clustering Coef.                 & 0.23  & 0.16  & 0.06   & 0.17 & 0.20 \\
        Avg. Neighbor Clustering Coef.        & 0.17  & 0.14  & 0.043  & 0.13 & 0.17 \\
        Number of Triangles                   & 1.88  & 1.53  & 1.90   & 176.93 & 2.28 \\
        Avg. PPRstd                           & 0.005 & 0.006 & 0.001  & 0.002 & 0.01 \\
        Diameter                              & 19    & 28    & 18     & 8 & 10 \\
        Number of Attributes                  & 1,433 & 3,703 & 500    & 50 & - \\
        Number of Classes                     & 7     & 6     & 3      & 121 & 4 \\
        Homophily                             & 0.80  & 0.73  & 0.80   & 0.88 & 0.79 \\
        Sparsity                              & 0.98  & 0.99  & 0.89   & 0.98 & - \\
    \end{tabular}
\end{minipage}
\hfill
\begin{minipage}{0.36\columnwidth}
    \centering
    \hspace*{-0.3cm}
    \includegraphics[width=0.25\linewidth]{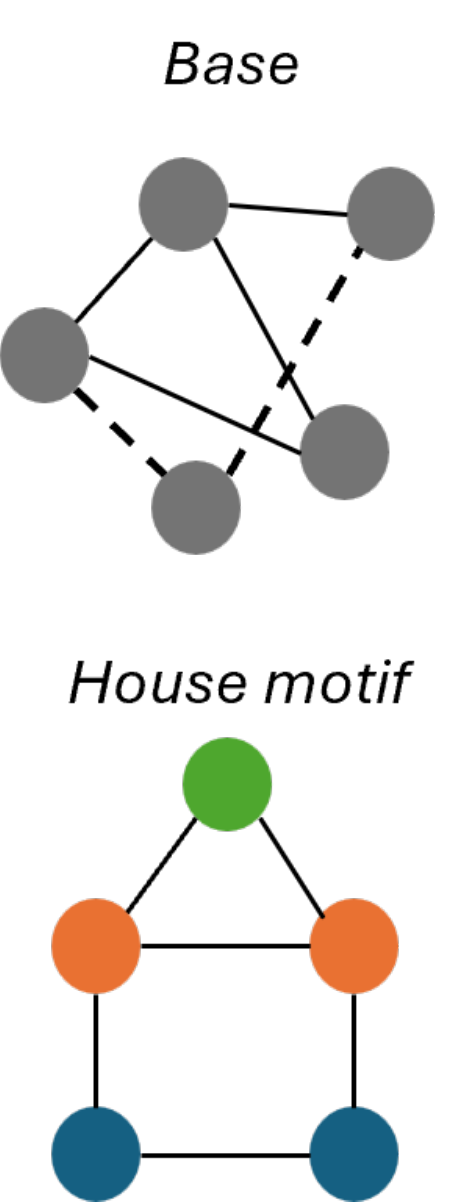}
\end{minipage}

\end{table}

In the unsupervised setting, we use node2vec \cite{grover2016node2vec}, role2vec \cite{ahmed2020role}, and three GNN encoders, GCN \cite{kipf2016semi}, GAT \cite{velickovic2017graph}, and GraphSAGE \cite{hamilton2017inductive}.
For the supervised setting, we train the same three GNN models. 
All models are implemented using pytorch geometric \cite{fey2019fast} and node representations are generated with 64 dimensions. More details on model training and performance can be found in the Appendix.

The interpretable model is trained using Linear Regression from the scikit-learn library \cite{pedregosa2011scikit}. This model is used as the default surrogate in our experiments. We additionally evaluated regularized linear variants, including Lasso and Ridge regression, as well as a nonlinear alternative based on HSIC Lasso. The results of these variants are presented in our ablation study. Across all experiments, we employ the weighted TACENR variant and select the affinity–divergence set ratio that achieves the best overall performance. For all experiments and the ablation study, we randomly sample a subset of nodes from each dataset and generate explanations along with their evaluation metrics. To obtain aggregated feature importances, we compute the mean importance score of each feature across the sampled nodes.

TACENR allows the incorporation of arbitrary sets of proximity and structural features, enabling the exploration of the importance of different features on different datasets and representation model architectures.
In our experiments we utilize a default set of structural features designed to capture complementary topological scopes. 
We use \textit{degree} to capture local first-order connectivity, while \textit{average neighbor degree} reflects second-order patterns by describing the connectivity of the immediate neighborhood of a node. To measure local density and motif participation, we incorporate \textit{clustering coefficient}, \textit{average neighbor clustering}, and \textit{node triangles}. Finally, to account for proximity-based influence beyond the local neighborhood, we include the standard deviation of personalized PageRank scores (\textit{PPRstd}). This set of features integrates measures of immediate connectivity volume, local community cohesiveness, and broader influence spread.
Table~\ref{tab:graph_stats} reports the average values of these features across all datasets. 
Experiments with different structural feature sets are presented in the ablation study.

\subsection{Explanations on Unsupervised Representation Methods}
To assess the quality and usefulness of TACENR explanations, we examine what information they reveal and whether these insights align with the architecture of the unsupervised representation models. We generate explanations for 10\% of randomly sampled nodes and analyze the ten features with the highest mean importance. Each explanation is reported together with the MSE of the interpretable model, which reflects the quality of the learned model.
We report results both on a synthetic dataset with ground-truth structural node characteristics and on the real-world datasets Cora, CiteSeer, PubMed, and PPI. 
 
\noindent \textbf{Explanations on the BA-Shapes synthetic dataset.}
To evaluate our explanation method, we conducted an experiment on the synthetic dataset BA-Shapes, which provides ground-truth structural node roles. This graph is designed to test whether models can distinguish global structural roles from local motif membership and is widely used for evaluating explanations in graph-based decision models. It consists of: (a) a base Barabási-Albert scale-free network, which is defined by a global power-law degree distribution, and (b) a set of house motifs which are dense, five-node subgraphs (houses) attached to the base graph.
Table~\ref{tab:graph_stats} illustrates both components: the base graph and the house motif. Within each house motif, nodes are categorized based on their structural position as bottom (blue), middle (orange), and top (green) nodes.

\begin{figure*}[htbp!]
    \centering  
    \includegraphics[width=\textwidth]{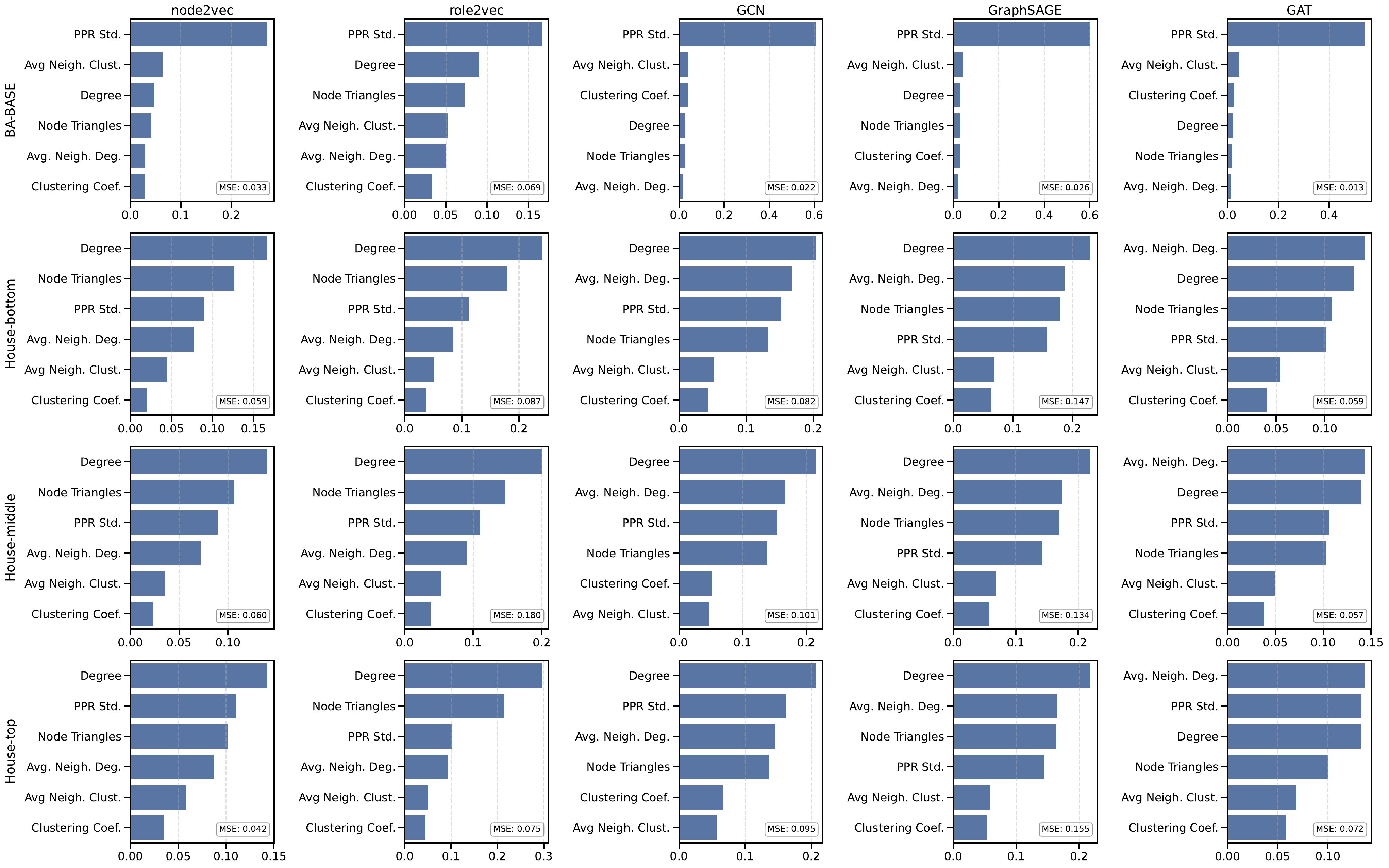}
    \caption{Feature importances for the BA-Shapes synthetic dataset across role2vec, node2vec, GCN, GAT, and GraphSAGE representation methods.}
    \label{fig:importances_BAShapes}
\end{figure*}

The goal of this experiment is to examine whether our TACENR explainer can correctly reveal that different representation learning models capture distinct structural rules for the different categories of nodes. We use role2vec which is known to encode structural characteristics of nodes into node representations, node2vec, and GNN encoders. The resulting feature importances, shown in Figure~\ref{fig:importances_BAShapes}, demonstrate that our explainer consistently identifies that all representation models learn two distinct structural rules corresponding to house-motif nodes and to non-motif (BA-BASE) nodes.

Across all representation models, the explanations for the three house node types are highly similar to each other but completely different from the BA-BASE nodes. 
For house-motif nodes, the dominant features are local structural features, such as \textit{degree}, \textit{node triangles}, and \textit{average neighbor clustering coefficient}. This is an expected result, as these nodes are defined by their local, dense motif structure. In contrast, for BA-BASE nodes, local features are shown to be less important. The explanations are almost exclusively dominated by the \textit{PPRstd} feature, which captures their global role as hubs or spokes in the scale-free backbone, rather than their immediate neighborhoods, which are not the defining factor for these nodes.

We also observe differences among the different representation models that reflect their architecture.  
For node2vec, a random-walk–based model, the most influential features for house nodes are \textit{degree}, \textit{node triangles}, and \textit{PPRstd}. This is expected, as random walks tend to remain within dense motifs, causing nodes in the same house to co-occur frequently and rely heavily on proximity signals.
For role2vec,  which is a model explicitly designed to capture structural roles, we see that \textit{degree} and \textit{node triangles} emerge as the most important features for house nodes.

Comparing the GNNs, we observe that for GCN, the most influential features are \textit{degree}, \textit{average neighbor degree}, and \textit{PPRstd}, reflecting its use of normalized multi-hop aggregation. The importance of  \textit{PPRstd} further aligns with the findings of \cite{nikolentzos2023gnns}, which show that GCN maps nodes to representations related to the sum of weighted normalized walks of length $k$.
For GraphSAGE, we find that structural features such as \textit{degree}, \textit{average neighbor degree}, and \textit{node triangles} are among the most influential. Through its neighbor sampling and aggregation mechanism, GraphSAGE captures patterns related to how well-connected and locally clustered the neighborhood of a node is.
For GAT, we observe that degree-related statistics consistently emerge as the most important features. This indicates that the attention mechanism prioritizes the connectivity patterns of neighboring nodes.

\begin{figure*}[htbp!]
    \centering
    \includegraphics[width=\textwidth]{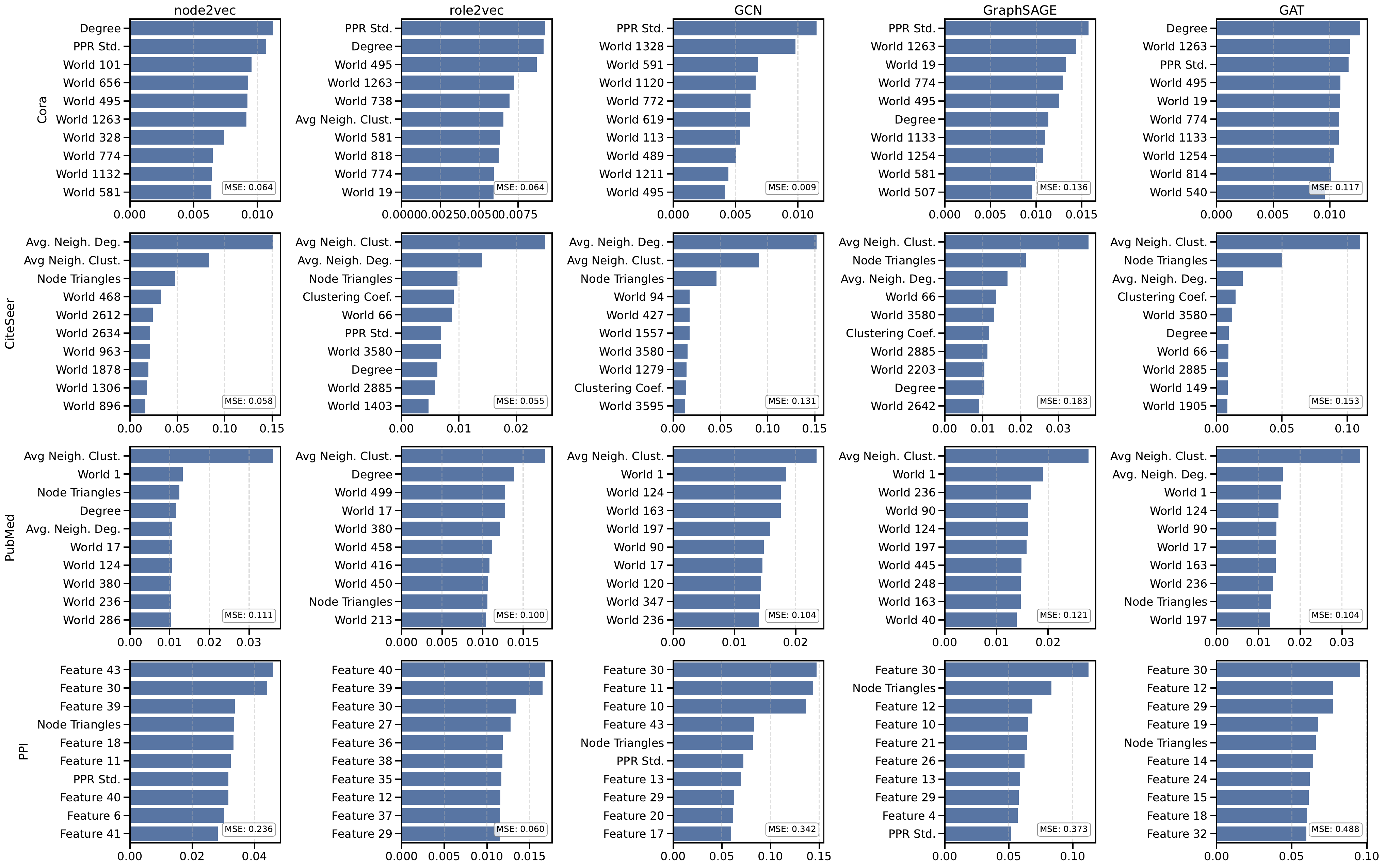}
    \caption{Feature importances across all real datasets and representation models.}
    \label{label:feature_importance_all_embeddings_no_proximity}
\end{figure*}

\noindent \textbf{Explanations on real datasets.}
We repeat this experiment on real-world datasets to assess how explanations behave in more complex, attribute-rich graphs. Figure \ref{label:feature_importance_all_embeddings_no_proximity} shows the explanations for node representations produced by node2vec, role2vec, GCN, GraphSAGE, and GAT.

Overall, we observe that structural features substantially influence the representation space in most datasets, with second-order structural features, such as \textit{average neighbor degree}, \textit{average neighbor clustering coefficient}, and \textit{node triangles} appearing among the most influential. The relative importance of structural, proximity, and attribute features varies across both models and datasets, reflecting the interaction between model architecture and graph topology.

Node2vec and role2vec assign higher importance to structural features. This is expected as the biased random walks of node2vec capture local co-occurrence patterns, while role2vec explicitly encodes structural roles. In contrast, GNN models yield very similar explanations within each dataset, suggesting that the structural properties of the dataset, rather than architectural differences, primarily determine which features shape the learned representations.

For Cora, we observe that the most important features are \textit{degree} and \textit{PPRStd} and several node attributes. This aligns with its high  homophily (0.8) and the relatively high average clustering coefficient (0.23), as shown in Table~\ref{tab:graph_stats}. High homophily makes attribute similarity important for the representation similarity, while the elevated clustering coefficient highlights the importance of local structural density captured by degree-based and proximity features.

For CiteSeer we observe that structural features such as \textit{average neighbor degree}, \textit{average neighbor clustering coefficient}, \textit{node triangles}, and \textit{clustering coefficient} rank among the most important features, surpassing node attributes. 
This behavior can be attributed to both the sparse graph topology (average clustering coefficient: 0.16) and the extreme attribute sparsity of the dataset (3,679 features with 0.99 sparsity), as shown in Table~\ref{tab:graph_stats}. In a graph where most nodes exist in simple, un-clustered neighborhoods, the few nodes that belong to a small motif or have a highly clustered neighborhood are structurally distinct. GNNs learn to exploit these structural patterns as highly discriminative.

For PubMed, the most influential features include \textit{average neighbor degree}, \textit{average neighbor clustering coefficient}, occasional \textit{node triangles}, and several  node attributes. Given the high homophily (0.8) of the dataset, moderate sparsity (0.89), and fewer features (500), attribute similarity contributes strongly to the learned representations.

For the PPI dataset, the explanations differ substantially from those of the citation networks. 
PPI is dense, high-degree, and triangle-rich (average degree 31.49, 170 triangles per node), making local structural features less discriminative.
Consequently, the representation models rely heavily on node attributes, which encode rich biological information, while \textit{node triangles} and occasionally \textit{PPRstd}, as a global diffusion-based measure, also contribute.  

\subsection{Comparison with Task-Specific Approaches}

Since no existing method explains the full node representation, we compare our supervised variant with four widely used task-specific explanation methods, GraphLIME \cite{huang2022graphlime}, GNNExplainer \cite{ying2019gnnexplainer}, GraphSVX \cite{duval2021graphsvx}, and COMPINEX \cite{giorgi2025combinex}, all of which return attribute-level importance scores. For fair comparison, we train a variant of TACENR that uses only node attributes as input to the interpretable model. 
For this comparison, we randomly sample 200 nodes and generate explanations. We compare methods using the Area Over the Perturbation Curve (AOPC), which evaluates whether the identified features are truly important, and by adding noisy features to test the ability of each method to ignore irrelevant information.


\noindent \textbf{Comparison using the AOPC curve.}
AOPC sequentially perturbs features based on their ranked importance and measures the resulting impact on the model performance. 
It is a standard evaluation metric for feature attribution methods and has been widely adopted in the evaluation of GNN explainers \cite{amara2022graphframex}. 
The AOPC score is calculated as the average cumulative drop in the predicted class probability up to the given rank, averaged across all instances. 
AOPC captures the necessity of an explanation, as higher AOPC values indicate that removing the features identified as important by the explainer leads to a larger change in the prediction of the model. 

\begin{figure*}[htbp!]
    \centering
    \includegraphics[width=\textwidth]{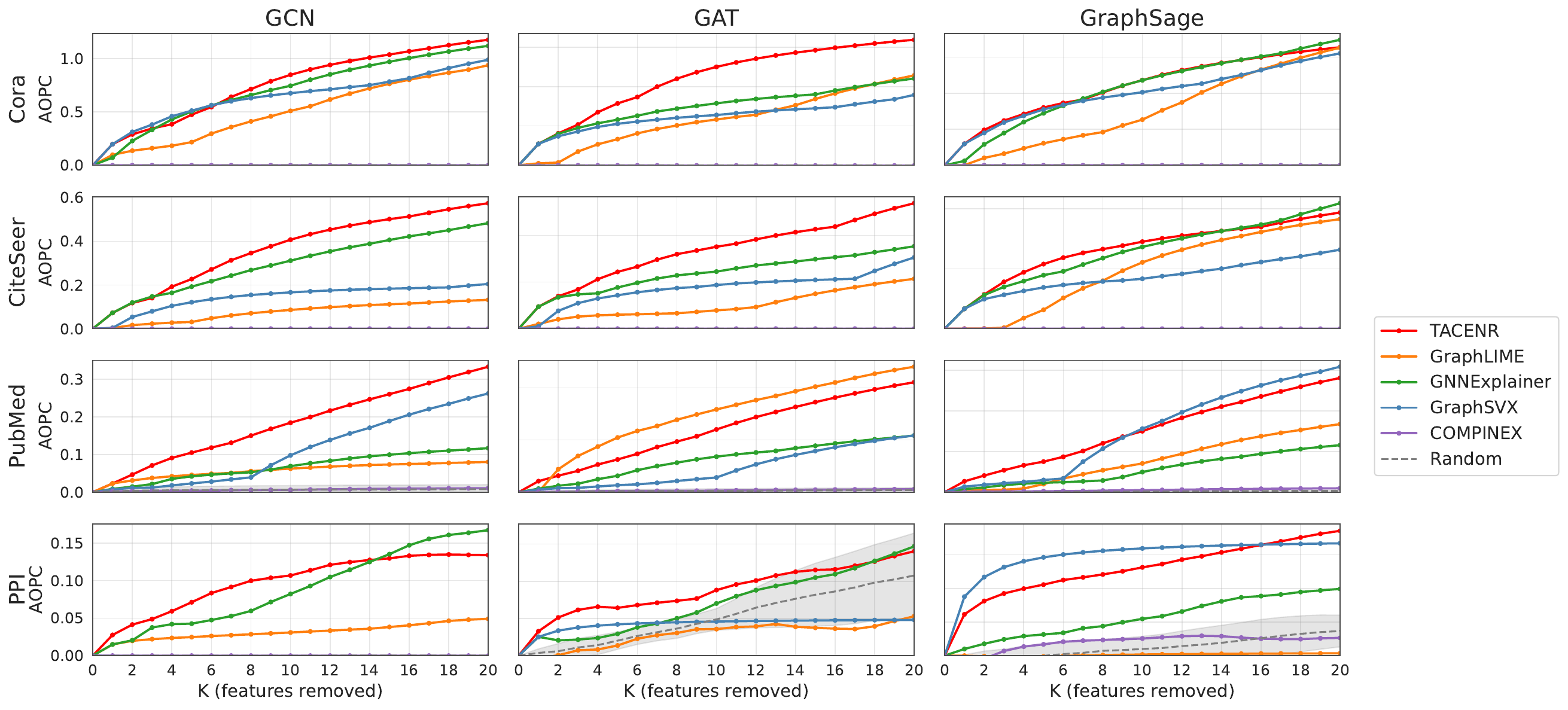}
    \caption{AOPC curves across all datasets for GCN, GAT, and GraphSAGE using TACENR, GraphLIME, and GNEExplainer explainers. }
    \label{fig:AOPC}
\end{figure*}
Figure \ref{fig:AOPC} shows the AOPC curves for all models and across all datasets. We also include a curve based on random ranking of features for comparison. A steeper and high curve indicates more effective feature ranking. We observe that all methods outperform the random baseline on most datasets. An exception is PPI, where, particularly for GAT, none of the methods achieve a clear separation from the random baseline but just for the first removed features. 
This is expected as PPI is different from the citation networks, with 121 classes and highly attribute-dominated representations, making it harder for explainers to identify a small set of important features.
Our method has the higher line in most cases, often together with GNEExplainer or GraphSVX, which also have a good performance. We also observe that COMPINEX often underperforms in this setting, likely because its explanations distribute importance across both node features and edges, reducing its ability to isolate the most influential node attributes for feature-only AOPC evaluation. 

\noindent \textbf{Comparison on filtering useless features.}
In this experiment, we evaluate the ability of our TACENR explainer to filter out irrelevant features, thus producing meaningful explanations. This type of evaluation is commonly used in explanation methods that return feature importance scores from interpretable or surrogate models \cite{huang2022graphlime,ribeiro2016should,duval2021graphsvx}, as 
it assess whether the features identified as important are indeed relevant.

We introduce noise by appending 10 randomly generated features to each node in Cora, CiteSeer, and PubMed, and 3 such features to nodes in PPI, which has far fewer original attributes. These noisy features are drawn from the same distribution as the real attributes, and we retrain the GNN on the augmented data.
We then generate explanations for the sampled nodes and evaluate each method based on how often it assigns importance to these noisy features.

\begin{figure*}[htbp!]
    \centering
    \includegraphics[width=\textwidth]{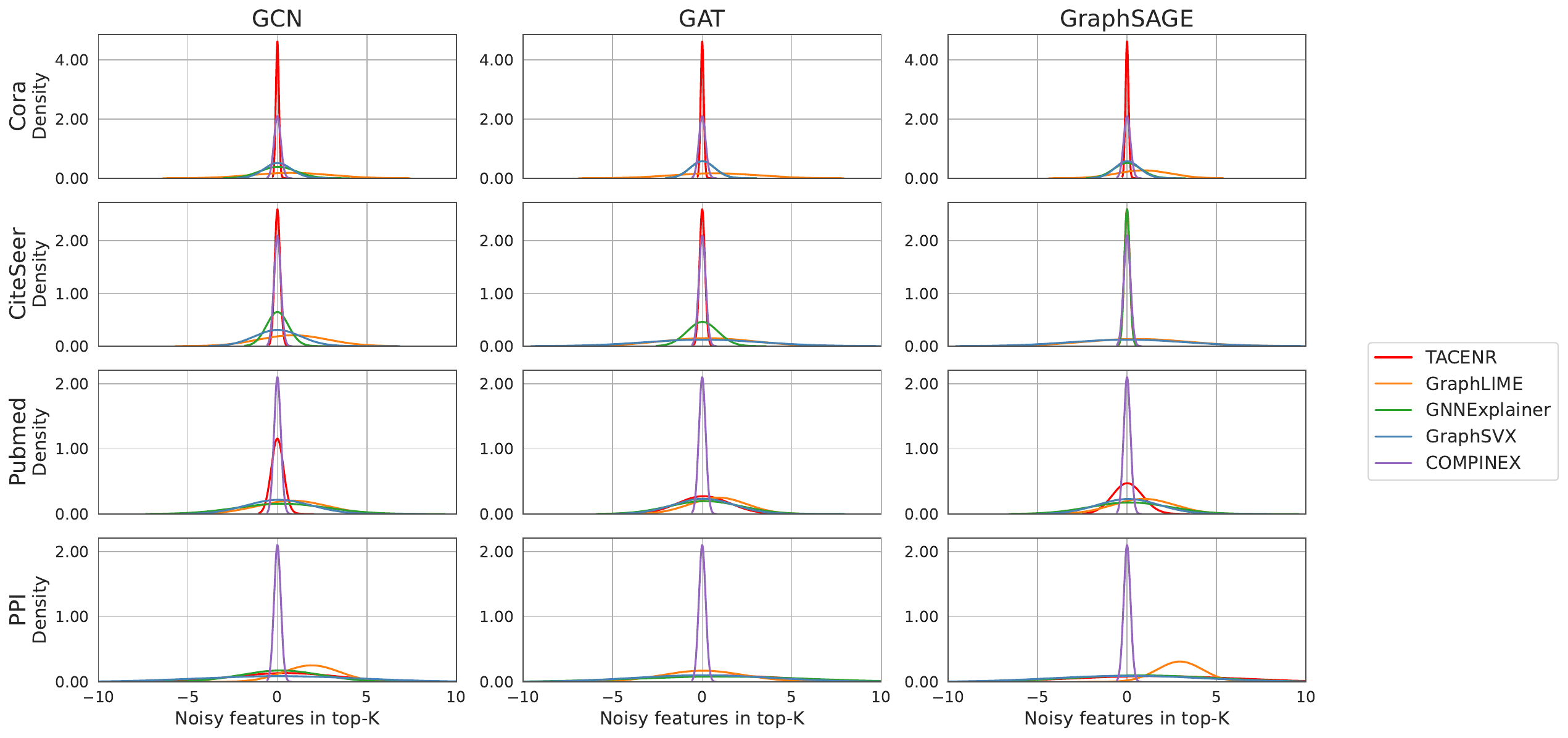}
    \caption{Distribution of noisy features in explanations.}
    \label{fig:noisy_features}
\end{figure*}
Figure \ref{fig:noisy_features} shows the frequency distributions of the number of noisy features selected by different explanation methods across all datasets and representation learning algorithms. These distributions were estimated using kernel density estimation (KDE) with a Gaussian kernel. We observe that our approach, together with COMPINEX, consistently selects fewer noisy features compared to the other methods, often showing a peak near zero noisy features. This indicates that our method rarely includes irrelevant or useless features in its explanations. In contrast, while the other methods also perform well, they tend to select more noisy features overall. Among them, GraphSVX and GNNExplainer exhibit relatively strong performance. The PPI dataset again displays distinct behavior, with most methods selecting noisy features more frequently, a pattern expected given that PPI contains far fewer features than the other datasets.

\subsection{Ablation Study}
In this section, we conduct an ablation study to examine several factors that influence the behavior and performance of TACENR. Specifically, we analyze the impact of structural feature selection, the effect of incorporating weighted similarity into the contrastive sampling process, the role of fine-tuning the affinity–divergence set composition, and the performance of different choices of interpretable surrogate models.

\subsubsection{Influence of structural feature selection.}
To analyze the impact of feature selection on TACENR explanations, we conduct an ablation study evaluating how different feature groups affect the resulting importance scores.

\begin{figure*}[htbp!]
    \centering
    \begin{subfigure}{\textwidth}
        \centering
        \includegraphics[width=\textwidth]{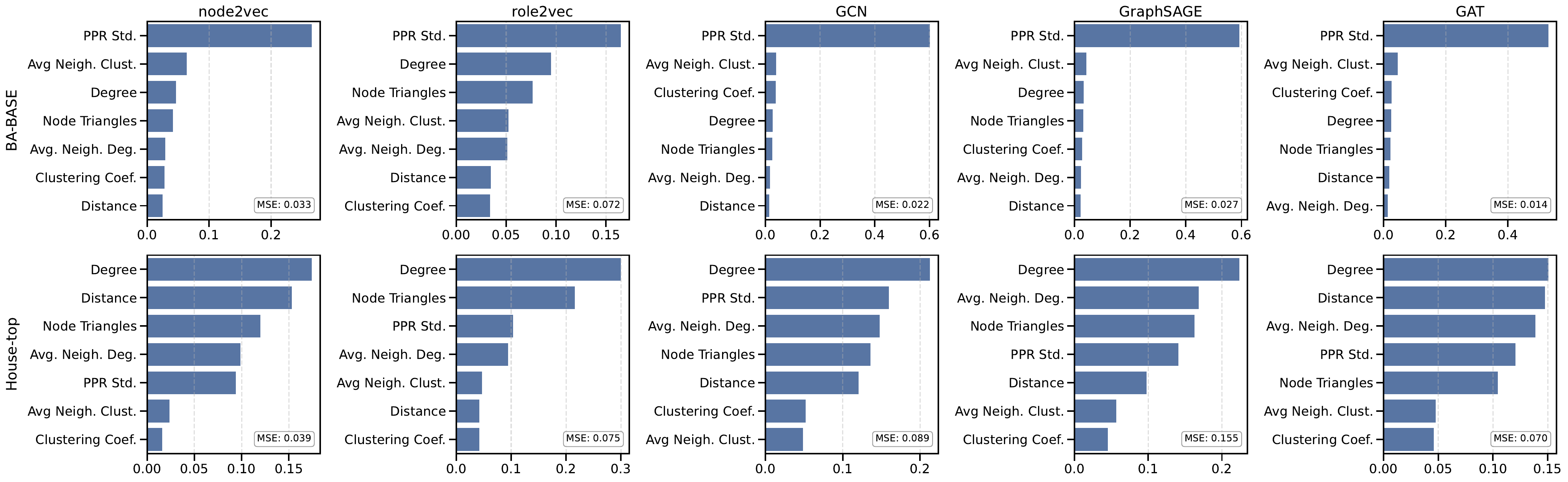}
        \caption{With the \textit{Distance} feature.}
        \label{fig:BA_Shapes_with_proximity}
    \end{subfigure}

    \begin{subfigure}{\textwidth}
        \centering
        \includegraphics[width=\textwidth]{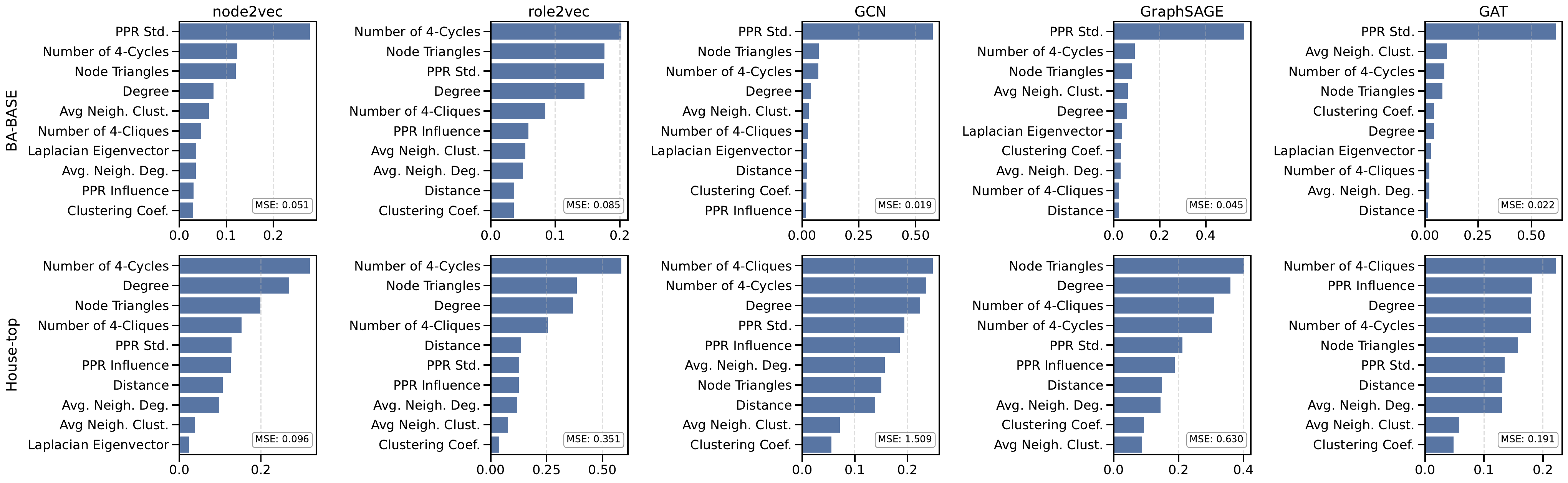}
        \caption{With full structural feature set.}
        \label{fig:BA_Shapes_all_features}
    \end{subfigure}

    \caption{Feature importances on the base and bottom nodes of BAShapes dataset across all representation learning models under two feature configurations.}
    \label{fig:BA_Shapes_compined}
\end{figure*}

\begin{figure*}[htbp!]
    \centering
    \begin{subfigure}{\textwidth}
        \centering
        \includegraphics[width=\textwidth]{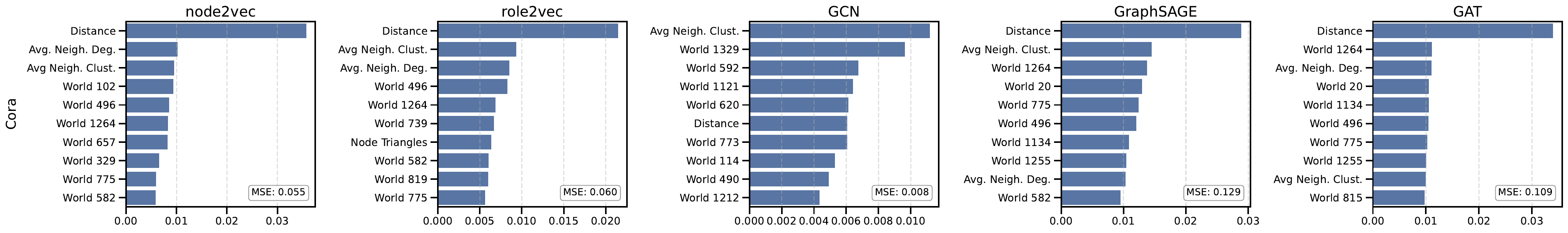}
        \caption{With the \textit{Distance} feature.}
        \label{fig:Cora_with_proximity}
    \end{subfigure}

    \begin{subfigure}{\textwidth}
        \centering
        \includegraphics[width=\textwidth]{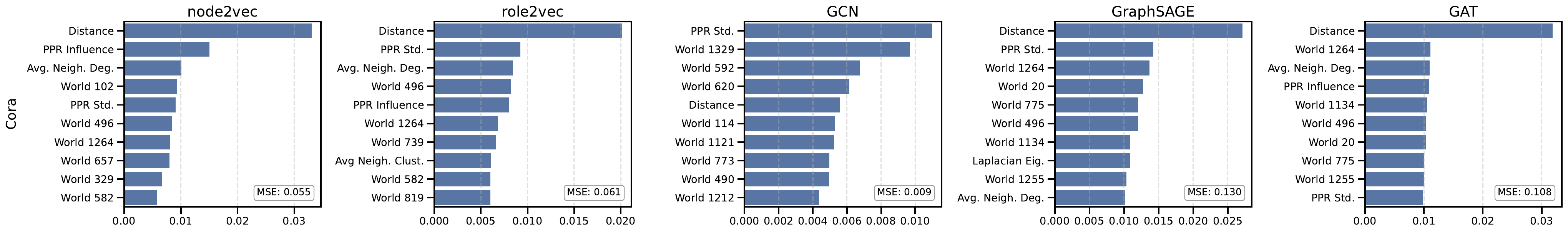}
        \caption{With full structural feature set.}
        \label{fig:Cora_all_features}
    \end{subfigure}

    \caption{Feature importances on the Cora dataset across all representation learning models under two feature configurations.}
    \label{fig:Cora_combined}
\end{figure*}

In this experiment, we introduce an additional proximity feature, \textit{distance}, that directly encodes topological similarity. Specifically, we define it as  $p_{v,u} = \frac{1}{1+dist(v ,u)}$
where $dist(v ,u)$ denotes the shortest path distance between nodes $v$ and $u$.
Figure~\ref{fig:BA_Shapes_with_proximity} shows the TACENR explanations for the BA-Shapes dataset after incorporating this feature, while Figure~\ref{fig:Cora_with_proximity} presents the corresponding results for the Cora dataset.

For the BA-Shapes dataset, we observe no substantial changes in the explanations, the representations continue to be driven primarily by local structural features, reflecting the motif-based nature of the graph.
In contrast, adding the \textit{distance} feature to Cora leads to a slight increase in explanation quality across all models, as evidenced by a decrease in MSE. Moreover, the \textit{distance} feature consistently emerges as one of the most influential features across all models.  
This indicates that the similarity between two nodes in the embedding space is closely related to their topological closeness within the graph.

We further experiment with an extended structural feature set by adding an additional proximity-based feature the \textit{PPR influence}, higher-order motif counts \textit{4-cliques} and \textit{4-cycles}, and a spectral descriptor, the normalized \textit{laplacian eigenvector}. Figure~\ref{fig:BA_Shapes_all_features} shows the resulting explanations for the BA-Shapes dataset, while Figure~\ref{fig:Cora_all_features} presents the corresponding results for the Cora dataset.

For BA-Shapes we observe that the higher-order motif features \textit{4-cliques} and \textit{4-cycles}
become particularly important across models. This is expected as BA-Shapes is explicitly constructed from repeated house motifs and a scale-free backbone, and its node identities are strongly determined by their role within these motif structures.

For Cora, as shown in Figure \ref{fig:Cora_all_features}, extending the feature set does not lead to a noticeable improvement in explanation quality, as measured by the MSE score. Across all representation learning models, \textit{distance} remains among the most influential features, while \textit{PPRstd} and \textit{PPR influence} also consistently emerge as highly ranked descriptors. We observe that higher-order motif features \textit{4-cliques} and \textit{4-cycles} seem not to be important. This behavior can be attributed to the structural characteristics of the Cora graph, which exhibits relatively small local neighborhoods, as indicated by its low average degree (Table~\ref{tab:graph_stats}).

Overall, the usefulness of proximity and structural features is highly dataset dependent. In motif-driven synthetic graphs such as BA-Shapes, higher-order structural descriptors such as \textit{4-cliques} and local connectivity measures play an important role. In contrast, for real-world attribute-rich graphs such as Cora, adding more structural features provides little additional benefit, explanations remain dominated by proximity-based descriptors and node attributes. TACENR enables experimentation with alternative feature sets, revealing which combinations influence the learned representations.

\subsubsection{Effect of weighted similarity on explanation quality.}
In this experiment, we explore whether incorporating dimension-wise weighting into the similarity metric improves explanation quality.
Weighted cosine similarity helps identify better affinity and divergence sets by emphasizing informative embedding dimensions.
However, we found that using the weighted similarity as the regression target degrades surrogate performance as it introduces additional variance across representation dimensions, making the regression task harder for a linear model.

To evaluate the impact of weighted similarity on contrastive set construction, we generate explanations for 10\% of randomly sampled Cora nodes. Tables \ref{tab:cora_error_scores_supervised} and \ref{tab:cora_error_scores_unsupervised} report MSE and MAE for all weighting variants, along with the mean improvement over the unweighted case. Statistical significance is assessed using paired t-tests, with $p < 0.05$ indicating a meaningful improvement.

In the supervised setting, we compute these weights based on the relevance of each dimension to the downstream task using gradient-based weighting and an alternative, Fisher SNR weights. 
Fisher SNR weights \cite{gu2011generalized} assign higher weights to representation dimensions that best discriminate between classes. For each dimension the weights are computed as the ratio of between class variance to within class variance. A higher value indicates that the corresponding dimension varies strongly across classes while remaining consistent within each class, making it more discriminative and thus more informative for similarity computation. 

Table~\ref{tab:cora_error_scores_supervised} shows that gradient-based weighting consistently improves both MSE and MAE, with statistically significant gains, indicating that emphasizing task-relevant dimensions enhances explanation quality. Fisher SNR weighting also improves performance, though not always significantly. 
Additional results for the remaining datasets are provided in the Appendix, where we observe that Fisher SNR weighting performs strongly on several datasets, making it a useful alternative when gradient signals are weak or unstable.
 
In the unsupervised setting, we compute the weights using the global variance of the dimensions and also an alternative where we use the difference between the global and local variance of each dimension. 
The underlying idea of global-local variance weighting is to identify dimensions whose local variance, computed over the $k_{aff}$ nearest neighbors of the node, deviates from the global variance, assigning higher importance to dimensions that exhibit distinctive local behavior.

\begin{table}[htbp!]
\caption{MSE and MAE on the Cora dataset for the supervised case, including percentage improvements and p-values after weighting.}
\centering
\tiny
\begin{tabular}{lcccccc}
\hline
\multicolumn{7}{c}{\textbf{Gradient-based weighting}} \\
\hline
\textbf{Model} 
& \textbf{MSE} & \textbf{MSE(W)} & \textbf{\%Imp.} 
& \textbf{MAE} & \textbf{MAE(W)} & \textbf{\%Imp.} \\
\hline
GCN       
& 0.113 & 0.102 & 9.11 ($<10^{-4}$)
& 0.237 & 0.225 & 4.92 ($<10^{-4}$) \\

GAT       
& 0.112 & 0.102 & 8.71 ($<10^{-4}$)
& 0.236 & 0.226 & 4.32 ($<10^{-4}$) \\

GraphSAGE 
& 0.008 & 0.006 & 24.71 ($<10^{-4}$)
& 0.065 & 0.056 & 14.38 ($<10^{-4}$) \\
\hline

\multicolumn{7}{c}{\textbf{Fisher SNR weighting}} \\
\hline
\textbf{Model} 
& \textbf{MSE} & \textbf{MSE(W)} & \textbf{\%Imp.} 
& \textbf{MAE} & \textbf{MAE(W)} & \textbf{\%Imp.} \\
\hline
GCN       
& 0.097 & 0.094 & 3.43 (0.07)
& 0.224 & 0.222 & 0.83 (0.37) \\

GAT       
& 0.222 & 0.218 &  1.91 (0.18)
& 0.351 & 0.346 &  1.46 (0.06)\\

GraphSAGE 
& 0.019 & 0.018 & 3.11 (0.07)
& 0.095 & 0.09 & 1.62 (0.04) \\
\hline
\end{tabular}
\label{tab:cora_error_scores_supervised}
\end{table}

Table~\ref{tab:cora_error_scores_unsupervised} summarizes the effect of global variance and global–local variance weighting on the quality of the interpretable model in the unsupervised case. This scenario is inherently more challenging than the supervised case, since there is no downstream task to directly indicate how informative each embedding dimension is.
Overall, both weighting strategies consistently improve MSE and MAE scores across most models compared to the unweighted baseline. The improvements are particularly notable for GCN and GAT: for example GCN, achieves a 9.59\% MSE improvement ($p < 10^{-4}$) under global variance weighting, and both GCN and GAT show highly significant gains (7.74\% and 6.81\%, respectively) with the global-local method. These results indicate that even in the absence of supervised signals, variance-based weighting effectively highlights structurally meaningful dimensions, thereby enhancing explanation quality.

\begin{table}[htbp!]
\caption{MSE and MAE on the Cora dataset for the unsupervised case, including percentage improvements and p-values after weighting.}
\centering
\tiny
\begin{tabular}{lcccccc}
\hline
\multicolumn{7}{c}{\textbf{Global variance weighting}} \\
\hline
\textbf{Model}
& \textbf{MSE} & \textbf{MSE(W)} & \textbf{\%Imp.}
& \textbf{MAE} & \textbf{MAE(W)} & \textbf{\%Imp.} \\
\hline
node2vec   & 0.053 & 0.052 & 2.09 (0.11)
           & 0.170 & 0.160 & 1.08 (0.14) \\
role2vec   & 0.061 & 0.060 & 1.55 (0.02)
           & 0.190 & 0.188 & 1.07 (0.007) \\
GCN        & 0.073 & 0.066 & 9.59 ($<10^{-4}$)
           & 0.195 & 0.186 & 4.70 ($<10^{-4}$) \\
GAT        & 0.048 & 0.046 & 2.97 (0.24)
           & 0.1549 & 0.154 & 0.13 (0.87) \\
GraphSAGE  & 0.071 & 0.070 & 1.10 (0.6)
           & 0.197 & 0.197 & 0 \\
\hline

\multicolumn{7}{c}{\textbf{Global--local variance weighting}} \\
\hline
node2vec   & 0.053 & 0.051 & 3.81 (0.02)
           & 0.171 & 0.167 & 1.99 (0.02) \\
role2vec   & 0.061 & 0.061 & 0
           & 0.189 & 0.189 & 0 \\
GCN        & 0.073 & 0.067 & 7.74 ($<10^{-4}$)
           & 0.195 & 0.187 & 4.01 ($<10^{-4}$) \\
GAT        & 0.048 & 0.045 & 6.81 ($<10^{-4}$)
           & 0.154 & 0.153 & 1.01 (0.22) \\
GraphSAGE  & 0.071 & 0.068 & 3.71 (0.05)
           & 0.197 & 0.192 & 2.33 (0.007) \\
\hline
\end{tabular}
\label{tab:cora_error_scores_unsupervised}
\end{table}

\subsubsection{Fine-tuning contrastive set composition.}

A core component of the architecture of our explanation method is the composition of the two contrastive sets, the affinity set $S^+_{v}$ and the divergence set $S^-_{v}$. To identify the most suitable balance between these sets, we fine-tune their composition to achieve the most accurate interpretable model.

In this experiment, for each of the sampled nodes, we construct its contrastive dataset by selecting 10\% of the nodes of the graph, which are used to train the corresponding interpretable model.  Within this training set, we vary the proportion of affinity and divergence nodes, evaluating five affinity–divergence ratios: 20\%/80\%, 40\%/60\%, 50\%/50\%, 60\%/40\%, and 80\%/20\%.
We then select the combination of affinity and divergence sets that achieves the best performance according to MSE and MAE metrics.

In supervised representations, explanation fidelity peaks when the surrogate model is trained on a set dominated by similar nodes (80\% affinity/20\% divergence). This can be attributed to the topological structure of supervised embeddings, which form tight, label-driven clusters, so a high-affinity sample effectively captures the dense intra-class features. 
Conversely, unsupervised representations require a dominance of dissimilar nodes to maximize performance (20\% affinity/80\% divergence). In unsupervised settings, structurally related nodes often possess highly correlated representations due to feature smoothing \cite{yang2020toward}, resulting in low variance within the positive set.
By increasing the proportion of dissimilar nodes, we provide the interpretable model with the necessary contrastive background to better learn the structural patterns that lead to similarity or dissimilarity in the embedding space.

\subsubsection{Comparison of interpretable models.} 
In this experiment, we evaluate the effectiveness of different types of models. In addition to standard Linear Regression, we consider two regularized linear models, Ridge and Lasso regression, as well as a nonlinear feature selection method based on HSIC Lasso \cite{yamada2014high}.
While Ridge and Lasso act as standalone predictive models, HSIC Lasso is specifically designed to identify nonlinear dependencies between features and targets and has also been used in GraphLIME \cite{huang2022graphlime}.
Since HSIC Lasso is a feature selector rather than a predictor, we employ it to identify the top-$k$ most relevant features. We set $k$ to 10\% of the available features, with an upper bound of 100 features. We then train a standard linear regression model using only this selected subset of features to generate predictions. 
For Ridge and Lasso regression, hyperparameters are selected via cross-validation.

\begin{table*}[htbp!]
\caption{MSE and MAE for all models, with significance reported relative to the Linear model.}
\label{tab:lasso_ridge_unsupervised}
\centering
\tiny
\begin{tabular}{ll|cc|cccc|cccc|cccc}
\hline
 & \textbf{Model}
 & \multicolumn{2}{c|}{\textbf{Linear}} 
 & \multicolumn{4}{c|}{\textbf{Lasso}} 
 & \multicolumn{4}{c|}{\textbf{Ridge}} 
 & \multicolumn{4}{c}{\textbf{HSIC Lasso}} \\
\cline{3-16}
 &
 & \textbf{MSE} & \textbf{MAE}
 & \textbf{MSE} & $\boldsymbol{p}$ & \textbf{MAE} & $\boldsymbol{p}$
 & \textbf{MSE} & $\boldsymbol{p}$ & \textbf{MAE} & $\boldsymbol{p}$
 & \textbf{MSE} & $\boldsymbol{p}$ & \textbf{MAE} & $\boldsymbol{p}$ \\
\hline

\multirow{5}{*}{Cora}
 & node2vec
 & 0.039 & 0.148
 & 0.036 & - & 0.143 & -
 & 0.048 & \ding{51} & 0.165 & \ding{51}
 & 0.041 & \ding{51} & 0.148 & \ding{51} \\
 & role2vec
 & 0.050 &  0.168
 & 0.055 & \ding{51} & 0.162 & -
 & 0.050 & - & 0.167 & -
 & 0.061 & \ding{51} & 0.175 & \ding{51}\\
 & GCN
 &  0.066 & 0.187
 & 0.065 & - & 0.175 & -
 & 0.061 & - & 0.180 & -
 & 0.007 & - & 0.059 & - \\
 & GAT
 & 0.047 & 0.163
 & 0.040 & - & 0.148 & -
 & 0.044 & - & 0.158 & -
 & 0.122 & \ding{51} & 0.253 &  \ding{51}\\
 & GraphSAGE
 & 0.070 & 0.198
 & 0.068 & - & 0.191 & -
 & 0.064 & - & 0.190 & -
 & 0.171 & \ding{51} & 0.286 & \ding{51} \\
\hline

\multirow{5}{*}{CiteSeer}
 & node2vec
 & 0.040 & 0.156
 & 0.040 & - & 0.151 & -
 & 0.043 & \ding{51} & 0.156 & -
 & 0.045 & \ding{51} & 0.159 & \ding{51} \\
 & role2vec
 & 0.074 & 0.200
 & 0.077 & \ding{51} & 0.203 & \ding{51}
 & 0.077 & \ding{51} & 0.202 & -
 & 0.060 & - & 0.178 & - \\
 & GCN
 & 0.048 & 0.150
 & 0.051 & \ding{51} & 0.144 & -
 & 0.046 & - & 0.151 & -
 & 0.133 & - & 0.256 & \ding{51}\\
 & GAT
 & 0.040 & 0.130
 & 0.041 & \ding{51} & 0.137 & \ding{51}
 & 0.036 & - & 0.137 & \ding{51}
 & 0.131 & - & 0.260 & \ding{51}\\
 & GraphSAGE
 & 0.050 & 0.160
 & 0.058 & \ding{51} & 0.164 & \ding{51}
 & 0.050 & - & 0.162 & \ding{51}
 & 0.183 & \ding{51} & 0.305 & \ding{51} \\
\hline

\multirow{5}{*}{PubMed}
 & node2vec
 & 0.097 & 0.223
 & 0.059 & - & 0.166 & -
 & 0.067 & - & 0.181 & -
 & 0.062 & - & 0.177 & - \\
 & role2vec
 & 0.038 & 0.137
 & 0.073 &\ding{51}  & 0.211 & \ding{51}
 & 0.074 & \ding{51} & 0.211 & \ding{51}
 & 0.084 & \ding{51} & 0.207 & \ding{51} \\
 & GCN
 & 0.077 & 0.200
 & 0.141 & \ding{51} & 0.207 & -
 & 0.085 & \ding{51} & 0.197 & -
 & 0.131 & \ding{51} & 0.235 & \ding{51} \\
 & GAT
 & 0.069 & 0.191
 & 0.099 & \ding{51} & 0.193 & -
 & 0.072 & \ding{51} & 0.185 & -
 & 0.090 & \ding{51} & 0.204 & \ding{51} \\
 & GraphSAGE
 & 0.097 & 0.223
 & 0.141 &\ding{51}  & 0.219 & -
 & 0.099 & - & 0.212 & -
 & 0.132 & \ding{51} & 0.239 & \ding{51} \\
\hline

\multirow{5}{*}{BA-Shapes}
 & node2vec
 & 0.023 & 0.111
 & 0.030 &\ding{51}  & 0.115 & -
 & 0.031 & \ding{51} & 0.115 & -
 & 0.037 &\ding{51}  & 0.121 & \ding{51} \\
 & role2vec
 & 0.045 & 0.163
 & 0.075 &\ding{51}  & 0.193 & \ding{51}
 & 0.074 & \ding{51} & 0.193 & \ding{51}
 & 0.062 & \ding{51} & 0.160 & - \\
 & GCN
 & 0.096 & 0.221
 & 0.106 & \ding{51} & 0.212 & -
 & 0.106 & \ding{51} & 0.212 & -
 & 0.047 & -- & 0.119 & - \\
 & GAT
 & 0.041 & 0.139
 & 0.039 & - & 0.131 & -
 & 0.039 & - & 0.131 & -
 & 0.057 & \ding{51} & 0.108 & - \\
 & GraphSAGE
 & 0.077 & 0.178
 & 0.120 & \ding{51} & 0.230 & \ding{51}
 & 0.122 & \ding{51} & 0.230 & \ding{51}
 & 0.081 & \ding{51} & 0.164 & - \\
\hline

\multirow{5}{*}{PPI}
 & node2vec
 & 0.237 & 0.362
 & 0.194 & - & 0.333 & -
 & 0.195 & - & 0.332 & -
 & 0.197 & - & 0.333 & - \\
 & role2vec
 & 0.060 &  0.176
 & 0.044 & - & 0.157 &- 
 & 0.044 & - & 0.157 & -
 &0.045  & - & 0.155 &-  \\
 & GCN
 & 0.342 & 0.409
 &  0.345& \ding{51} & 0.416 & \ding{51}
 & 0.343 & \ding{51} & 0.415 &\ding{51}
 &  0.450 & \ding{51} &  0.503 & \ding{51} \\
 & GAT
 & 0.862 & 0.670
 & 0.698 &-  & 0.648  & -
 & 0.696 & - & 0.644 & -
 &  0.707& - & 0.644 & - \\
 & GraphSAGE
 & 0.410 & 0.458
 & 0.398 & - & 0.462 & \ding{51}
 &0.397 & - & 0.461 & \ding{51}
 & 0.410 & - & 0.485 & \ding{51} \\
\hline

\end{tabular}
\end{table*}

Table~\ref{tab:lasso_ridge_unsupervised} presents the MSE and MAE scores, followed by statistical significance test, for the unsupervised case, we observe comparable results among models and in many cases simple Linear Regression achieves slightly better performance than the rest. The tick marks indicate cases where Linear Regression performs significantly better than the corresponding method according to the statistical significance test.
We observe similar results in the supervised case, where the performance of the Linear, Ridge, Lasso, and HSIC Lasso models remains comparable across most representation learning methods. 

Overall, as model performance is comparable across both supervised and unsupervised settings, we adopt plain Linear Regression as the default interpretable model, as it offers maximal interpretability through explicit coefficients for all features.

\section{Conclusion}
\label{conclusion}

In this work we present TACENR, a task-agnostic explanation method designed to interpret node representations in graph representation learning.
Unlike existing approaches that focus on task-specific explanations or explaining individual representation dimensions, TACENR explains the entire node representation.
To achieve this it trains an interpretable model contrastively to learn a similarity function in the representation space. 
It effectively identifies the features that most influence the position of node representation among proximity, structural and node attribute features.
TACENR is applicable to both supervised and unsupervised representations. In the contrastive learning process, we introduce a weighted similarity function that adapts to the type of representation. In supervised settings, the weighting emphasizes dimensions most correlated with the predicted label whereas for unsupervised representations, it prioritizes dimensions with the highest variance.
Our experimental results show that TACENR produces meaningful explanations that are consistent with the underlying structure of the representation learning models. When compared against existing task-specific explanation methods, using the AOPC metric and noisy features, our supervised variant consistently outperforms these methods in identifying the most important features.

\begin{credits}
\subsubsection{\ackname} This work has been partially supported by project MIS 5154714 of the National Recovery and Resilience Plan Greece 2.0 funded by the European Union under the NextGenerationEU Program.
\subsubsection{\discintname}
The authors have no competing interests to declare that are relevant to the content of this article.
\end{credits}

\bibliographystyle{splncs04}
\bibliography{mybibliography}

\appendix
\section*{APPENDIX}
This appendix provides additional details on our experimental configuration and further evaluations of the weighting-based similarity methods.

\section{Experimental Setup}
All experiments are conducted on a Linux machine with an NVIDIA RTX A6000 GPU (driver 565.57.01, CUDA 12.7). 
For supervised experiments, we use three GNN architectures (GCN, GAT, and GraphSAGE) with two message-passing layers and a linear output layer, producing 64-dimensional embeddings. 
Hyperparameters, including hidden size, dropout, learning rate, and weight decay, are tuned via random search. 
Models are trained using the Adam optimizer with early stopping based on validation performance. 
Table~\ref{tab:model_performance_supervised_noisy_features} reports training, validation, and test accuracies for all models and datasets on both the original and noise-augmented data.
 \begin{table*}[htbp!]
\centering
\caption{Performance of supervised models across datasets with original and noisy features.}
\label{tab:model_performance_combined_clean}
\tiny
\begin{tabular}{llcccccccccc}
\toprule
 & & \multicolumn{2}{c}{Cora} 
   & \multicolumn{2}{c}{CiteSeer} 
   & \multicolumn{2}{c}{PubMed}
   & \multicolumn{2}{c}{PPI} \\
\cmidrule(lr){3-4} 
\cmidrule(lr){5-6} 
\cmidrule(lr){7-8}
\cmidrule(lr){9-10}
Model & Split 
      & Clean & Noisy 
      & Clean & Noisy 
      & Clean & Noisy
      & Clean & Noisy \\
\midrule

\multirow{3}{*}{GCN}
 & Train & 1.00 & 1.00 & 1.00 & 0.95 & 1.00 & 1.00 &  0.88 & 0.83 \\
 & Val   & 0.81 & 0.81 & 0.69 & 0.71 & 0.81 & 0.80 &  0.86&  0.72\\
 & Test  & 0.82 & 0.81 & 0.67 & 0.70 & 0.76 & 0.77 &0.84  & 0.70 \\
\midrule

\multirow{3}{*}{GAT}
 & Train & 0.99 & 0.97 & 0.94 & 1.00 & 0.98 & 1.00 & 0.87 &0.87  \\
 & Val   & 0.82 & 0.81 & 0.71 & 0.71 & 0.80 & 0.80 & 0.88 & 0.88 \\
 & Test  & 0.81 & 0.82 & 0.70 & 0.69 & 0.77 & 0.77 & 0.87 & 0.89 \\
\midrule

\multirow{3}{*}{GraphSAGE}
 & Train & 1.00 & 1.00 & 1.00 & 0.96 & 1.00 & 1.00 & 0.87 & 0.82 \\
 & Val   & 0.80 & 0.78 & 0.69 & 0.71 & 0.80 & 0.72 & 0.87 & 0.88 \\
 & Test  & 0.79 & 0.79 & 0.68 & 0.69 & 0.76 & 0.70 & 0.84 & 0.71 \\
\bottomrule
\label{tab:model_performance_supervised_noisy_features}
\end{tabular}
\end{table*}
For unsupervised models, we use node2vec, role2vec, and GAE architectures with GCN, GraphSAGE, or GAT encoders, each producing 64-dimensional embeddings with two message-passing layers. 
node2vec is trained using biased random walks, with $p$ and $q$ tuned via grid search to maximize validation AUC. 
For autoencoder-based models, hyperparameters such as learning rate, dropout, and weight decay are tuned with early stopping based on validation performance. 
All models are trained using the Adam optimizer. 
Model quality is evaluated on a link prediction task using AUC and Average Precision (AP) on validation and test splits. 
Table~\ref{tab:model_performance_unsupervised} summarizes the results. 
\begin{table}
\centering
\caption{Performance of unsupervised models across datasets.}
\tiny
\begin{tabular}{llcccccccccccc}
\toprule
 & & 
 \multicolumn{2}{c}{Cora} & 
 \multicolumn{2}{c}{CiteSeer} & 
 \multicolumn{2}{c}{PubMed} &
 \multicolumn{2}{c}{PPI} &
 \multicolumn{2}{c}{BA-Shapes} \\
\cmidrule(lr){3-4} 
\cmidrule(lr){5-6} 
\cmidrule(lr){7-8}
\cmidrule(lr){9-10}
\cmidrule(lr){11-12}
 & & AUC & AP & AUC & AP & AUC & AP & AUC & AP & AUC & AP \\
\midrule

\multirow{3}{*}{node2vec} 
 & Val   & 0.87 & 0.89 & 0.82 & 0.86 & 0.82 & 0.87 & 0.87  &  0.89  & 0.86 & 0.87 \\
 & Test  & 0.86 & 0.89 & 0.77 & 0.82 & 0.82 & 0.87 & 0.87  & 0.89   & 0.87 & 0.87 \\
\midrule

\multirow{3}{*}{GCN} 
 & Val   & 0.92 & 0.93 & 0.93 & 0.93 & 0.96 & 0.96 & 0.84  & 0.85 & 0.88 & 0.88\\
 & Test  & 0.93 & 0.94 & 0.92 & 0.93 & 0.96 & 0.96 & 0.85  &  0.85& 0.82 & 0.82 \\
\midrule

\multirow{3}{*}{GAT} 
 & Val   & 0.92 & 0.92 & 0.95 & 0.96 & 0.94 & 0.94 & 0.78  & 0.77  & 0.87 & 0.87 \\
 & Test  & 0.92 & 0.93 & 0.94 & 0.94 & 0.94 & 0.94 & 0.78  & 0.77  & 0.87 & 0.85 \\
\midrule

\multirow{3}{*}{GraphSAGE} 
 & Val   & 0.89 & 0.90 & 0.90 & 0.91 & 0.89 & 0.90 & 0.83  &  0.84 & 0.86 & 0.84 \\
 & Test  & 0.90 & 0.91 & 0.90 & 0.90 & 0.90 & 0.91 & 0.82  & 0.83  & 0.85 & 0.85 \\
\bottomrule

\end{tabular}
\label{tab:model_performance_unsupervised}
\end{table}

\section{Effect of Weighted Similarity on Explanation Quality.}

In this section, we present additional results comparing the weighting methods across multiple datasets: CiteSeer, Pubmed, and PPI for supervised representations, and CiteSeer, Pubmed, PPI, and BA-Shapes for unsupervised representations. Table \ref{tab:weights_supervised} present the results for the supervised case.

We observe that across all datasets, incorporating weighting into the similarity metric generally improves explanation quality. Gradient-based weighting produces consistent improvements in MSE and MAE across all datasets and GNN architectures, often with statistically significant gains. Fisher SNR weighting also seems to improve performance in most cases.
For unsupervised representations both the global-variance and global-local methods provides stable improvements, especially on CiteSeer, PubMed and PPI. However, weighting does not help on BA-Shapes. BA-Shapes has easy to learn structural patterns, so reweighting the embedding dimensions cannot add useful information.

\begin{table*}[htbp!]
\caption{{\footnotesize MSE and MAE on CiteSeer, PubMed, and PPI datasets for the supervised case, including percentage improvements and p-values.}}
\centering
\tiny
\begin{tabular}{llcccccc}
\hline
\multicolumn{8}{c}{\textbf{Gradient-based weighting}} \\
\hline
 & \textbf{Model} 
& \textbf{MSE} & \textbf{MSE(W)} & \textbf{\%Imp.} 
& \textbf{MAE} & \textbf{MAE(W)} & \textbf{\%Imp.} \\
\hline

\multirow{3}{*}{CiteSeer}
 & GCN       & 0.050 & 0.040 & 20.51 ($<10^{-4}$)
              & 0.151 & 0.134 & 11.35 ($<10^{-4}$) \\
 & GAT       & 0.087 & 0.086 & 1.39 ($<10^{-4}$)
              & 0.200 & 0.200 & 0 \\
 & GraphSAGE & 0.008 & 0.005 & 40.13 ($<10^{-4}$)
              & 0.059 & 0.047 & 20.58 ($<10^{-4}$) \\
\hline

\multirow{3}{*}{PubMed}
 & GCN       & 0.118 & 0.123 & 0 
              & 0.240 & 0.245 & 0 \\
 & GAT       & 0.181 & 0.177 & 2.60 ($<10^{-4}$)
              & 0.296 & 0.293 & 1.11 \\
 & GraphSAGE & 0.017 & 0.015 & 8.40 ($<10^{-4}$)
              & 0.089 & 0.084 & 5.34 ($<10^{-4}$) \\
\hline

\multirow{3}{*}{PPI}
 & GCN       & 0.0018 & 0.001 & 37.7 ($<10^{-4}$)
              & 0.0248 & 0.024 &0.56 (0.6) \\
 & GAT       &0.115& 0.082 & 28.02 ($<10^{-4}$)
              & 0.236 & 0.21 & 10.83 ($<10^{-4}$) \\
 & GraphSAGE & 0.0003 & 0.0002 &44.11 ($<10^{-4}$)
              &0.008& 0.007 & 18.24 ($<10^{-4}$)  \\
\hline

\multicolumn{8}{c}{\textbf{Fisher SNR weighting}} \\
\hline
 & \textbf{Model} 
& \textbf{MSE} & \textbf{MSE(W)} & \textbf{\%Imp.} 
& \textbf{MAE} & \textbf{MAE(W)} & \textbf{\%Imp.} \\
\hline

\multirow{3}{*}{CiteSeer}
 & GCN       & 0.050 & 0.009 & 82.06 ($<10^{-4}$) & 0.151 & 0.063 & 58.19 ($<10^{-4}$)\\
 & GAT       & 0.087 & 0.09 & 0  & 0.200 & 0.209 & 0 \\
 & GraphSAGE & 0.008 & 0.01 & 0 & 0.059 & 0.065 & 0 \\
\hline

\multirow{3}{*}{PubMed}
 & GCN       & 0.118 & 0.115 & 2.45 (0.23) & 0.240 & 0.237 & 1.18 (0.12) \\
 & GAT       & 0.181 & 0.12 & 32.53 ($<10^{-4}$) & 0.296 & 0.241 & 18.46 ($<10^{-4}$)\\
 & GraphSAGE & 0.017 & 0.015 & 10.78 ($<10^{-4}$) & 0.089 & 0.085 & 4.61 ($<10^{-4}$)\\
\hline

\multirow{3}{*}{PPI}

 & GCN       & 0.0018 & 0.001 & 38.83 ($<10^{-4}$)
              &0.024 & 0.02 & 3.18 ($<10^{-4}$)\\
 & GAT       & 0.115& 0.085 & 25.68 ($<10^{-4}$)
              & 0.236 & 0.2 & 14.93 ($<10^{-4}$) \\
 & GraphSAGE & 0.0003 & 0.0002 & 40.33 ($<10^{-4}$)
              & 0.0088 & 0.008 & 3.56 (0.008) \\
\hline

\end{tabular}
\label{tab:weights_supervised}
\end{table*}

\begin{table*}[ht]
\caption{{\footnotesize MSE and MAE on CiteSeer, PubMed, PPI, and BA-Shapes datasets for the unsupervised case, including percentage improvements and p-values.}}
\centering
\tiny
\begin{tabular}{llcccccc}
\hline
\multicolumn{8}{c}{\textbf{Global Variance Weighting}} \\
\hline
 & \textbf{Model}
& \textbf{MSE} & \textbf{MSE(W)} & \textbf{\%Imp.}
& \textbf{MAE} & \textbf{MAE(W)} & \textbf{\%Imp.} \\
\hline
\multirow{5}{*}{CiteSeer}
 & node2vec   & 0.053 & 0.050 & 5.42 (0.29)
              & 0.164 & 0.162 & 1.69 (0.35) \\
 & role2vec   & 0.080 & 0.074 & 7.16 ($<10^{-4}$)
              & 0.204 & 0.200 & 1.83 ($<10^{-4}$) \\
 & GCN        & 0.056 & 0.048 & 14.55 (0.01)
              & 0.162 & 0.150 & 7.32 (0.003) \\
 & GAT        & 0.040 & 0.040 & 0
              & 0.130 & 0.130 & 0 \\
 & GraphSAGE  & 0.050 & 0.050 & 0
              & 0.160 & 0.160 & 0 \\
\hline
\multirow{5}{*}{PubMed}
 & node2vec   & 0.114 & 0.097 & 14.52 ($<10^{-4}$)
              & 0.247 & 0.223 & 9.69 ($<10^{-4}$) \\
 & role2vec   & 0.038 & 0.038 & 0
              & 0.137 & 0.137 & 0 \\
 & GCN        & 0.093 & 0.077 & 17.08 ($<10^{-4}$)
              & 0.222 & 0.200 & 9.80 ($<10^{-4}$) \\
 & GAT        & 0.082 & 0.069 & 15.69 ($<10^{-4}$)
              & 0.212 & 0.191 & 9.92 ($<10^{-4}$) \\
 & GraphSAGE  & 0.114 & 0.097 & 14.52 ($<10^{-4}$)
              & 0.247 & 0.223 & 9.69 ($<10^{-4}$) \\
\hline
\multirow{5}{*}{PPI}
 & node2vec   & 0.210  & 0.200  & 4.49 (0.04)  & 0.317  & 0.311  & 1.94 (0.01)  \\
 & role2vec   & 0.061  & 0.049  & 18.50 ($<10^{-4}$)  & 0.177  & 0.159  & 10.12 ($<10^{-4}$)  \\
 & GCN        & 0.363  & 0.244  & 32.67 ($<10^{-4}$)  & 0.408  & 0.352  &  13.73 ($<10^{-4}$)\\
 & GAT        &  0.464 & 0.421  &  9.41 (0.04) &  0.454 &  0.487 &  0 \\
 & GraphSAGE  & 0.380  &  0.228 &40.01 ($<10^{-4}$)   & 0.447  &  0.349 &  21.88 ($<10^{-4}$)\\
\hline
\multirow{5}{*}{BA-Shapes}
 & node2vec   & 0.035 & 0.032 & 8.83 (0.26)  & 0.118 & 0.123 & 3.76 \\
 & role2vec   & 0.048 & 0.047 & 0             & 0.166 & 0.166 & 0 \\        
 & GCN        & 0.100 & 0.095 & 4.74 (0.36)   & 0.210 & 0.205 & 2.32 (0.29) \\        
 & GAT        & 0.033 & 0.031 & 5.18 (0.49)   & 0.112 & 0.108 & 3.49 (0.31) \\
 & GraphSAGE  & 0.073 & 0.085 & 0             & 0.153 & 0.178 & 0 \\
             
\hline
\multicolumn{8}{c}{\textbf{Global--Local Variance Weighting}} \\
\hline
 & \textbf{Model}
& \textbf{MSE} & \textbf{MSE(W)} & \textbf{\%Imp.}
& \textbf{MAE} & \textbf{MAE(W)} & \textbf{\%Imp.} \\
\hline
\multirow{5}{*}{CiteSeer}
 & node2vec   & 0.053 & 0.050 & 5.07 (0.42)
              & 0.164 & 0.161 & 2.19 (0.34) \\
 & role2vec   & 0.080 & 0.071 & 11.52 ($<10^{-4}$)
              & 0.204 & 0.198 & 2.41 (0.027) \\
 & GCN        & 0.056 & 0.051 & 8.69 (0.004)
              & 0.162 & 0.156 & 3.79 (0.006) \\
 & GAT        & 0.040 & 0.040 & 0
              & 0.139 & 0.144 & 0 \\
 & GraphSAGE  & 0.053 & 0.054 & 0
              & 0.160 & 0.164 & 0 \\
\hline
\multirow{5}{*}{PubMed}
 & node2vec   & 0.071 & 0.093 & 0
              & 0.200 & 0.220 & 0 \\
 & role2vec   & 0.038 & 0.038 & 0
              & 0.130 & 0.140 & 0 \\
 & GCN        & 0.093 & 0.078 & 16.27 ($<10^{-4}$)
              & 0.220 & 0.201 & 9.45 ($<10^{-4}$) \\
 & GAT        & 0.082 & 0.069 & 15.53 ($<10^{-4}$)
              & 0.212 & 0.192 & 9.62 ($<10^{-4}$) \\
 & GraphSAGE  & 0.114 & 0.098 & 13.84 ($<10^{-4}$)
              & 0.247 & 0.224 & 9.11 ($<10^{-4}$) \\
\hline
\multirow{5}{*}{PPI}
 & node2vec   &  0.241 &  0.195 &  18.86 ($<10^{-4}$) &  0.366 & 0.311  & 15.14 ($<10^{-4}$)  \\
 & role2vec   &  0.061 & 0.042  & 30.40 ($<10^{-4}$)  & 0.177  & 0.150  & 14.92 ($<10^{-4}$)  \\
 & GCN        & 0.363  & 0.245  & 32.46 ($<10^{-4}$)  & 0.408  & 0.358  & 12.19 ($<10^{-4}$)   \\
 & GAT        &  0.464  & 0.433  &  6.68 (0.03) & 0.454  & 0.493  & 0  \\
 & GraphSAGE  &  0.380 &  0.229 & 39.67 ($<10^{-4}$)  & 0.447  & 0.351  &  21.54 ($<10^{-4}$) \\
\hline
\multirow{5}{*}{BA-Shapes}
 & node2vec   & 0.025 & 0.023 & 10.28 (0.16) & 0.109 & 0.111 & 0 \\
 & role2vec   & 0.048 & 0.045 & 5.95 (0.17)  & 0.166 & 0.163 & 1.66 (0.49) \\        
 & GCN        & 0.100 & 0.096 & 3.45 (0.61)  & 0.220 & 0.210 & 0 \\        
 & GAT        & 0.033 & 0.041 & 0            & 0.112 & 0.139 & 0 \\
 & GraphSAGE  & 0.073 & 0.077 & 0            & 0.153 & 0.178 & 0 \\
             
\hline
\end{tabular}
\label{tab:weights_unsupervised}
\end{table*}

\end{document}